# Takagi–Sugeno Fuzzy Modeling and Control for Effective Robotic Manipulator Motion

**Izzat Al-Darraji[1,2], Ayad A. Kakei[2], Ayad Ghany Ismaeel[3], Georgios Tsaramirsis[4], Fazal Qudus Khan[5], Princy Randhawa[6], Muath Alrammal[4] and Sadeeq Jan[7,*]**

[1]Department of Automated Manufacturing, University of Baghdad, Baghdad, 10001, Iraq
[2]Department of Mechanical Engineering, University of Kirkuk, Kirkuk, 36001, Iraq
[3]Department of Computer Technical Engineering, Al-kitab University, Kirkuk, 36001, Iraq
[4]Higher Colleges of Technology, Abu Dhabi Women's College, Abu Dhabi, 41012, UAE
[5]Department of Computer Science, University of Swat, Shangla Campus, Alpurai, 19100, Shangla, Pakistan
[6]Department of Mechatronics Engineering, Manipal University Jaipur, Jaipur, 302004, India
[7]Department of Computer Science & IT, University of Engineering & Technology Peshawar, Peshawar, 25000, Pakistan
*Corresponding Author: Sadeeq Jan. Email: sadeeqjan@uetpeshawar.edu.pk
Received: 06 August 2021; Accepted: 07 September 2021

**Abstract:** Robotic manipulators are widely used in applications that require fast and precise motion. Such devices, however, are prompt to nonlinear control issues due to the flexibility in joints and the friction in the motors within the dynamics of their rigid part. To address these issues, the Linear Matrix Inequalities (LMIs) and Parallel Distributed Compensation (PDC) approaches are implemented in the Takagy–Sugeno Fuzzy Model (T-SFM). We propose the following methodology; initially, the state space equations of the nonlinear manipulator model are derived. Next, a Takagy–Sugeno Fuzzy Model (T-SFM) technique is used for linearizing the state space equations of the nonlinear manipulator. The T-SFM controller is developed using the Parallel Distributed Compensation (PDC) method. The prime concept of the designed controller is to compensate for all the fuzzy rules. Furthermore, the Linear Matrix Inequalities (LMIs) are applied to generate adequate cases to ensure stability and control. Convex programming methods are applied to solve the developed LMIs problems. Simulations developed for the proposed model show that the proposed controller stabilized the system with zero tracking error in less than 1.5 s.

## 1 Introduction

Recently, robots have been applied widely in situations that require precise movement at high speeds. In this type of robotic application, the flexibilities in joints [1] and friction [2] are critical factors of the modeling and control processes. To provide precise tracking despite of the existing high nonlinearity effects of joint flexibilities and motor friction, advanced control techniques are essential to be taken in the control design stage. Generally, nonlinearity is an essential issue that was and is still the focus of researchers [3–7]. Techniques such as backstepping control [8], impedance control [9], and sliding mode control [10] are some of the methods used in the control stage. The main issue of these techniques is that the high order nonlinearity for fast motion robotics cannot be solved. In contrast, Takagy–Sugeno Fuzzy Model (T-SFM) is an effective method for representing the high order nonlinear systems in terms of a combination of state equations [11–13]. In this study, initially, the nonlinear state space representation of the robot system is derived. It is assumed that the high nonlinearities of 3rd order in the equation of spring torque of the flexible joint and friction model of Striebeck effect. These nonlinear sources in the state space equations are substituted by rules of T-SFM. The T-SFM is implemented to linearize the derived state space equation. A controller is designed from the developed T-SFM using the method of Parallel Distributed Compensation (PDC). The prime concept of the designed controller is to deduce all the fuzzy rules to compensate for all rules of the fuzzy model. Furthermore, the linear matrix inequalities (LMIs) are applied to generate adequate cases for approving the stability and control purpose issues. Convex programming methods are applied to solve the developed LMIs problems.

This paper is organized as follows. In Section 2, the related work of this study is presented. In Section 3, the state and output equations of the robot arm are derived. In Section 4, the nonlinear robot arm model is represented by the Takagi–sugeno model. Section 5 describes the designe of the controller for the nonlinear robot system. In Section 6, the results of this paper are verified through simulation tests. Finally, the conclusions are written in Section 7.

## 2 Related Work

Nonlinearity is an essential issue that has been considered by various researchers in the control of robot manipulators. In [14], T-SFM is implemented with a sliding mode controller to solve the issues of the systems having nonlinear dynamic behavior. This proposed controller showed the high efficiency of avoiding chattering within very accurate tracking. In [15], the nonlinear parts in the dynamic system equation are identified by using T-SFM model. Next, using the obtained model from the T-SFM technique, a new controller is designed depending on estimation principle, *i.e.*, not whole measurements. It is shown that nonlinear partial differential equation systems can be controlled using an incomplete number of actuators and sensors. In [16], T-SFM with decomposition controller approach are applied to control the desired trajectory of aircraft with taking into consideration existing both the inaccuracies in the aircraft model and disturbances. The introduced control technique stabilized the closed loop system and tracked in an asymptotically, for a reference step input signal, the pitch angle of the aircraft. In [17], T-SFM approach is applied to represent a partly-active chair suspension system of electrorheological damper. The application of the T-SFM approach has simplified the design process of the H∞ controller. Compared to the existing control technique, the introduced Takagy–Sugeno control technique improved the performance of the the electrorheological damper partly active chair suspension system. In [18], T-SFMs are applied in the robust control of non-constant speed wind turbines which utilizes a generator of twice-fed induction type. The suggested Takagy–Sugeno control method provides the optimum power under considering non-constant wind speed. In [19], an adaptive T-SFM is proposed for piezoelectric actuators to overcome the issue of nonlinearity behavior due to the hysteresis features which degrades the tracking performance. The presented T-SFM showed its efficiency in controlling the piezoelectric actuators without needing the mathematical representation of the hysteresis model. Furthermore, the values of the T-SFM are tuned online to handle the errors of tracking. In [20], an adaptive T-SFM is implemented in the design of a permanent magnetic generator that uses a turbine system. The adaptive T-SFM is assured of the ability to supply electricity in a robust and reliable way. In [21], T-SFM is applied to control an omnidirectional ball robot manipulator of motors which is equipped on two orthogonal planes. The proposed T-SFM control method satisfied high performance results. In which, the model rules corresponding to zero and five degrees limited the omnidirectionally ball robot manipulator performance by narrowing the control range. In [22], TS-FM technique is applied with observers in commercial vehicles to accomplish the appropriate sensors that are necessary to realize precise torque sensing. The rule of the TS-FM was to treat the increasing of nonlinearities in the driving as opposed to load when the speed of the vehicle increased. Based on the above features of applying T-SFM in various applications, this study is focused on the issues of nonlinearities due to joint flexibilities and motor friction consideration for fast and precise motions robot manipulator by T-SFM control.

## 3 Development of State Space Model

In this section, the state and output equations of the robot arm model are developed. Selection of the appropriate components [23] and modeling is the important step that should be implemented before any development [24,25]. Firstly, equations of motion of the robot arm shown in Fig. 1 are derived. The robot arm model has three main parts: motor, gear, and robot arm [26]. The inertia of the motor, gear and robot arm are denoted by $J_{motor}$, $J_{gear}$, and $J_{arm}$, respectively. The robot arm system is actuated by the input torque $\tau_{in}$ of the motor. Besides this, the friction effect is assumed to act on the motor within a nonlinear torsion coefficient $K_{f,motor}$. Due to the flexibility effect, the input torque rotates the motor, the gear and robot arm within an unequal angular position $\theta_{motor}$, $\theta_{gear}$, and $\theta_{arm}$, respectively. The flexibility of the gearbox is modeled by a nonlinear torsion stiffness $K_{gb}$. The gear ratio of the gearbox is assumed one, $i.e.$, $n_{gear} = 1$. Since this study focused on the nonlinearity of the joints, the flexibility in the robot arm is modeled by a linear spring of torsion stiffness $K_{arm}$. Besides the effect of flexibility in the gearbox and the robot arm, the damping coefficients in the gearbox and the robot arm are considered $d_{gb}$ and $d_{arm}$, respectively.

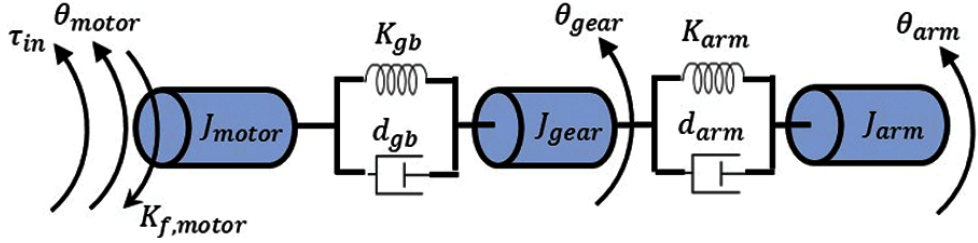

**Figure 1:** The model of the nonlinear robot arm

The nonlinear torque friction part is considered as the effect of coulomb friction and Striebeck effect. The coulomb friction is assumed as [27].

$$K_{C,f,motor} = \mu N \dot{\theta}_{motor}, \tag{1}$$

The total nonlinear torque friction effect is considered as Tustin friction model [28]:

$$K_{f,motor} = \left( K_{C,f,motor} + \left( K_{S,f,motor} - K_{C,f,motor} \right) e^{\frac{\dot{\theta}_{motor}}{v_s}} \right) sign(\dot{\theta}_{motor}) + K_{v,f,motor} \dot{\theta}_{motor} \tag{2}$$

where $K_{S,f,motor}$, $v_s$, and $K_{v,f,motor}$ denote static friction, striebeck velocity, and viscous friction, respectively. On the other hand, the nonlinear torsion torque of gear model is assumed as

$$\tau_{gb} = k_{gb,1} \left( \dot{\theta}_{motor} - \dot{\theta}_{gear} \right) + k_{gb,2} (\dot{\theta}_{motor} - \dot{\theta}_{gear})^3 \tag{3}$$

Applying newton second law, the equation of motion for the motor inertia, gear inertia, and arm inertia is

$$J_{motor}\ddot{\theta}_{motor} = -K_{gb}(\theta_{motor} - \theta_{gear}) - d_{gb}\left( \dot{\theta}_{motor} - \dot{\theta}_{gear} \right) - K_{f,motor}\dot{\theta}_{motor} + \tau_{in}, \tag{4}$$

$$J_{gear}\ddot{\theta}_{gear} = K_{gb}\left( \theta_{motor} - \theta_{gear} \right) + d_{gb}\left( \dot{\theta}_{motor} - \dot{\theta}_{gear} \right) - K_{arm}\left( \theta_{gear} - \theta_{arm} \right) - d_{arm}\left( \dot{\theta}_{gear} - \dot{\theta}_{arm} \right), \tag{5}$$

$$J_{arm}\ddot{\theta}_{arm} = K_{arm}\left( \theta_{gear} - \theta_{arm} \right) + d_{arm}\left( \dot{\theta}_{gear} - \dot{\theta}_{arm} \right), \tag{6}$$

Respectively. To get the state space model, assume the following state variables: $x_1 = \theta_{motor}$, $x_2 = \theta_{gear}$, $x_3 = \theta_{arm}$, $x_4 = \dot{\theta}_{motor}$, $x_5 = \dot{\theta}_{gear}$, $x_6 = \dot{\theta}_{arm}$, and the output is the angular velocity of the motor that is assumed $y = x_4$. thus, the differential states are,

$$\dot{x}_1 = x_4, \tag{7}$$
$$\dot{x}_2 = x_5, \tag{8}$$
$$\dot{x}_3 = x_6, \tag{9}$$
$$\dot{x}_4 = \frac{1}{J_{motor}} \left( -K_{gb} \left( x_1 - x_2 \right) - d_{gb} \left( x_4 - x_5 \right) - K_{f,motor} x_4 + \tau_{in} \right), \tag{10}$$
$$\dot{x}_5 = \frac{1}{J_{gear}} \left( K_{gb} \left( x_1 - x_2 \right) + d_{gb} \left( x_4 - x_5 \right) - K_{arm} \left( x_2 - x_3 \right) - d_{arm} \left( x_5 - x_6 \right) \right), \tag{11}$$
$$\dot{x}_6 = \frac{1}{J_{arm}} \left( K_{arm} \left( x_2 - x_3 \right) + d_{arm} \left( x_5 - x_6 \right) \right), \tag{12}$$

The differential states from Eqs. (7)–(12) and the output formula, $i.e.$, the angular velocity of the motor, can be arranged in state space model format as

$$\begin{bmatrix} \dot{x}_1 \\ \dot{x}_2 \\ \dot{x}_3 \\ \dot{x}_4 \\ \dot{x}_5 \\ \dot{x}_6 \end{bmatrix} = A\left(t\right) \begin{bmatrix} x_1 \\ x_2 \\ x_3 \\ x_4 \\ x_5 \\ x_6 \end{bmatrix} + B\left(t\right), \tag{13}$$

$$y = C\left(t\right),$$

where

$$A\left(t\right) = \begin{bmatrix} 0 & 0 & 0 & 1 & 0 & 0 \\ 0 & 0 & 0 & 0 & 1 & 0 \\ 0 & 0 & 0 & 0 & 0 & 1 \\ \frac{-K_{gb}}{J_{motor}} & \frac{K_{gb}}{J_{motor}} & 0 & \frac{-d_{gb}-K_{f,motor}}{J_{motor}} & \frac{d_{gb}}{J_{motor}} & 0 \\ \frac{K_{gb}}{J_{gear}} & \frac{-K_{gb}-K_{arm}}{J_{gear}} & \frac{K_{arm}}{J_{gear}} & \frac{d_{gb}}{J_{gear}} & \frac{-d_{gb}-d_{arm}}{J_{gear}} & \frac{d_{arm}}{J_{gear}} \\ 0 & \frac{K_{arm}}{J_{arm}} & \frac{-K_{arm}}{J_{arm}} & 0 & \frac{d_{arm}}{J_{arm}} & \frac{-d_{arm}}{J_{arm}} \end{bmatrix},$$

$$B\left(t\right) = \begin{bmatrix} 0 \\ 0 \\ 0 \\ \tau_{in} \\ 0 \\ 0 \end{bmatrix},$$

$$C\left(t\right) = x_4.$$

The obtained state model in Eq. (13) will be implemented in the next section to be represented by the T-SFM approach. The numerical results of this study are based on the physical parameters of the robot model presented in Tab. 1.

**Table 1:** Proposed robot manipulator parameters

| Physical parameter | Value (Unit) |
|---|---|
| $J_{motor}$ | $6.3 \times 10^{-3} \ (kg.m^2)$ |
| $J_{gear}$ | $40 \times 10^{-3} \ (kg \cdot m^2)$ |
| $J_{arm}$ | $13 \times 10^{-3} \ (kg \cdot m^2)$ |
| $K_{arm}$ | $6 \ (N/m)$ |
| $d_{gb}$ | $35 \times 10^{-3} \ (N \cdot s/m)$ |
| $d_{arm}$ | $95 \times 10^{-3} \ (N \cdot s/m)$ |

After deriving the state equations of our proposed nonlinear robot model, T-SFM can be implemented to linearize the robot system as explained in the next section.

## 4 Implementation of T-SFM

T-SFM is implemented to represent the nonlinear state space model of the robot arm. In addition, the T-SFM has linearized the nonlinear terms in state space equations. The nonlinear robot arm model will be denoted by T-SFM as

$$\dot{x}(t) = \sum_{j=1}^{r} \omega_j(z(t))(A_j x(t) + B_j u(t) + a_j), \tag{14}$$

$$y(t) = \sum_{j=1}^{r} \omega_j(z(t))(C_j x(t) + c_i), \tag{15}$$

where $r, Z(t)$, and $\omega_j(z(t))$, and "$a_j$ & $c_j$" denotes the number of local models, scheduling variables vector, normalized membership function, and the biases of the $j$th local model, respectively. Within T-SFM, the nonlinear robot arm model is exemplified in a close combination of state variables. Consider the range of the state variables as follow: $x_1(t) \in [0, 2\pi]$, $x_2(t) \in [0, 2\pi]$, $x_3(t) \in [0, 2\pi]$, $x_4(t) \in [0, 10]$, $x_5(t) \in [0, 10]$, and $x_6(t) \in [0, 10]$. In Eq. (13), the nonlinear elements due to nonlinear properties of $K_{f,motor}$ and $K_{gb}$ are assumed: $z_1(t) = \frac{-K_{gb}}{J_{motor}}$, $z_2(t) = \frac{K_{gb}}{J_{motor}}$, $z_3(t) = \frac{-d_{gb} - K_{f,motor}}{J_{motor}}$, and $z_4(t) = \frac{-K_{gb} - K_{arm}}{J_{gear}}$. Inserting these formulas in the value of matrix $A$ of Eq. (13), results:

$$A(t) = \begin{bmatrix} 0 & 0 & 0 & 1 & 0 & 0 \\ 0 & 0 & 0 & 0 & 1 & 0 \\ 0 & 0 & 0 & 0 & 0 & 1 \\ z_1(t) & z_2(t) & 0 & z_3(t) & \frac{d_{gb}}{J_{motor}} & 0 \\ z_2(t) & z_4(t) & \frac{K_{arm}}{J_{gear}} & \frac{d_{gb}}{J_{gear}} & \frac{-d_{gb} - d_{arm}}{J_{gear}} & \frac{d_{arm}}{J_{gear}} \\ 0 & \frac{K_{arm}}{J_{arm}} & \frac{-K_{arm}}{J_{arm}} & 0 & \frac{d_{arm}}{J_{arm}} & \frac{-d_{arm}}{J_{arm}} \end{bmatrix}, \tag{16}$$

Under our considered range values, the minimum and maximum values of $z_1(t)$, $z_2(t)$, $z_3(t)$, $z_4(t)$ are calculated considering the values of the parameters which are involved in their formulas. In terms of Eq. (14), the scheduling variables can be represented now as

$$z_1(t) = L_1(z_1(t)).\text{Max } z_1(t) + L_2(z_1(t)).\text{Min } z_1(t), \tag{17}$$

$$z_2(t) = M_1(z_2(t)).\text{Max } z_2(t) + M_2(z_1(t)).\text{Min } z_2(t), \tag{18}$$

$$z_3(t) = N_1(z_3(t)).\text{Max } z_3(t) + N_2(z_3(t)).\text{Min } z_3(t), \tag{19}$$

$$z_4(t) = T_1(z_4(t)).\text{Max } z_4(t) + T_2(z_4(t)).\text{Min } z_4(t) \tag{20}$$

where $L_1(z_1(t)) + L_2(z_1(t)) = 1$, $M_1(z_2(t)) + M_2(z_1(t)) = 1$, $N_1(z_3(t)) + N_2(z_3(t)) = 1$, and $T_1(z_4(t)) + T_2(z_4(t)) = 1$. Consequently, the membership functions are obtained as:

$$L_1 = \frac{z_1(t) - \text{Min } z_1(t)}{z_1(t).\text{Max } z_1(t) - z_1(t).\text{Min } z_1(t)}, \tag{21}$$

$$L_2 = \frac{z_1(t) - \text{Max } z_1(t)}{z_1(t).\text{Min } z_1(t) - z_1(t).\text{Max } z_1(t)}, \tag{22}$$

$$M_1 = \frac{z_2(t) - \text{Min } z_2(t)}{z_2(t).\text{Max } z_2(t) - z_2(t).\text{Min } z_2(t)}, \tag{23}$$

$$M_2 = \frac{z_2(t) - \text{Max } z_2(t)}{z_2(t).\text{Min } z_2(t) - z_2(t).\text{Max } z_2(t)}, \tag{24}$$

$$N_1 = \frac{z_3(t) - \text{Min } z_3(t)}{z_3(t).\text{Max } z_3(t) - z_3(t).\text{Min } z_3(t)}, \tag{25}$$

$$N_2 = \frac{z_3(t) - \text{Max } z_3(t)}{z_3(t).\text{Min } z_3(t) - z_3(t).\text{Max } z_3(t)}, \tag{26}$$

$$T_1 = \frac{z_4(t) - \text{Min } z_4(t)}{z_4(t).\text{Max } z_4(t) - z_4(t).\text{Min } z_4(t)}, \tag{27}$$

$$T_2 = \frac{z_4(t) - \text{Max } z_4(t)}{z_4(t).\text{Min } z_4(t) - z_4(t).\text{Max } z_4(t)}, \tag{28}$$

these membership functions $L_1$, $L_2$, $M_1$, $M_2$, $N_1$, $N_2$, $T_1$, and $T_2$ are named as Big_1, Small_1, Big_2, Small_2, Big_3, Small_3, Big_4, Small_4, respectively.

## 5 Controller Design

The presented design of the controller is closely related to the feature of the derived T-SFM in the previous section. This feature is that the terms of state equations $z_1(t)$, $z_2(t)$, $z_3(t)$, and $z_4(t)$ are not constant. These terms are varying corresponding to the torsion torque $K_{gb}$ and motor friction $K_{f,motor}$ depends on the velocities of motor and gear of the state variables. This property was the source of the high nonlinearity in the robot arm model. Consequently, for linearization purpose, the nonlinear robot arm system is expressed by the bellow fuzzy model:

**Model rule 1**

If $z_1(t)$ is Small_1 and $z_2(t)$ is Small_2 and $z_3(t)$ is Small_3 and $z_4(t)$ is Small_4 then

$$\dot{x}(t) = A_1 x(t) + B_1 u(t)$$

**Model rule 2**

If $z_1(t)$ is Small_1 and $z_2(t)$ is Small_2 and $z_3(t)$ is Small_3 and $z_4(t)$ is Big_4 then

$$\dot{x}(t) = A_2 x(t) + B_2 u(t)$$

**Model rule 3**

If $z_1(t)$ is Small_1 and $z_2(t)$ is Small_2 and $z_3(t)$ is Big_3 and $z_4(t)$ is Small_4 then

$$\dot{x}(t) = A_3 x(t) + B_3 u(t)$$

**Model rule 4**

If $z_1(t)$ is Small_1 and $z_2(t)$ is Small_2 and $z_3(t)$ is Big_3 and $z_4(t)$ is Big_4 then

$$\dot{x}(t) = A_4 x(t) + B_4 u(t)$$

**Model rule 5**

If $z_1(t)$ is Small_1 and $z_2(t)$ is Big_2 and $z_3(t)$ is Small_3 and $z_4(t)$ is Small_4 then

$$\dot{x}(t) = A_5 x(t) + B_5 u(t)$$

**Model rule 6**

If $z_1(t)$ is Small_1 and $z_2(t)$ is Big_2 and $z_3(t)$ is Small_3 and $z_4(t)$ is Big_4 then

$$\dot{x}(t) = A_6 x(t) + B_6 u(t)$$

**Model rule 7**

If $z_1(t)$ is Small_1 and $z_2(t)$ is Big_2 and $z_3(t)$ is Big_3 and $z_4(t)$ is Small_4 then

$$\dot{x}(t) = A_7 x(t) + B_7 u(t)$$

**Model rule 8**

If $z_1(t)$ is Small_1 and $z_2(t)$ is Big_2 and $z_3(t)$ is Big_3 and $z_4(t)$ is Big_4 then

$$\dot{x}(t) = A_8 x(t) + B_8 u(t)$$

**Model rule 9**

If $z_1(t)$ is Big_1 and $z_2(t)$ is Small_2 and $z_3(t)$ is Small_3 and $z_4(t)$ is Small_4 then

$$\dot{x}(t) = A_9 x(t) + B_9 u(t)$$

**Model rule 10**

If $z_1(t)$ is Big_1 and $z_2(t)$ is Small_2 and $z_3(t)$ is Small_3 and $z_4(t)$ is Big_4 then

$$\dot{x}(t) = A_{10} x(t) + B_{10} u(t)$$

**Model rule 11**

If $z_1(t)$ is Big_1 and $z_2(t)$ is Small_2 and $z_3(t)$ is Big_3 and $z_4(t)$ is Small_4 then

$$\dot{x}(t) = A_{11} x(t) + B_{11} u(t)$$

**Model rule 12**

If $z_1(t)$ is Big_1 and $z_2(t)$ is Small_2 and $z_3(t)$ is Big_3 and $z_4(t)$ is Big_4 then

$$\dot{x}(t) = A_{12} x(t) + B_{12} u(t)$$

**Model rule 13**

If $z_1(t)$ is Big_1 and $z_2(t)$ is Big_2 and $z_3(t)$ is Small_3 and $z_4(t)$ is Small_4 then

$$\dot{x}(t) = A_{13} x(t) + B_{13} u(t)$$

**Model rule 14**

If $z_1(t)$ is Big_1 and $z_2(t)$ is Big_2 and $z_3(t)$ is Small_3 and $z_4(t)$ is Big_4 then

$$\dot{x}(t) = A_{14} x(t) + B_{14} u(t)$$

**Model rule 15**

If $z_1(t)$ is Big_1 and $z_2(t)$ is Big_2 and $z_3(t)$ is Big_3 and $z_4(t)$ is Small_4 then

$$\dot{x}(t) = A_{15} x(t) + B_{15} u(t)$$

**Model rule 16**

If $z_1(t)$ is Big_1 and $z_2(t)$ is Big_2 and $z_3(t)$ is Big_3 and $z_4(t)$ is Small_4 then

$$\dot{x}(t) = A_{16} x(t) + B_{16} u(t)$$

In this way, the nonlinear robot model can be described in the following general formula

$$Model\ rule\ j : If\ z_1(t)\ is\ M_1^j\ and\ z_2(t)\ is\ M_2^j\ and\ z_3(t)\ is\ M_3^j\ and\ z_4(t)\ is\ M_4^j \tag{29}$$
$$Then\ \dot{x}(t) = A_j x(t) + B_j u(t), \quad j = 1, \ldots, 16$$

Applying the defuzzification principle of fuzzy method, the output of the system is obtained as:

$$\dot{x}(t) = \frac{\sum_{j=1}^{r} \omega_j(z(t))(A_j x(t) + B_j u(t) + a_j)}{\sum_{j=1}^{r} \omega_j(z(t))} \tag{30}$$

$$y(t) = \frac{\sum_{j=1}^{r} \omega_j(z(t))(C_j x(t) + c_i)}{\sum_{j=1}^{r} \omega_j(z(t))} \tag{31}$$

where $\omega_j(z(t))$ denotes the membership function for the model rule $j$. Hence

$$\omega_j(z(t)) = \prod_{v=1}^{r} M_v^j(x_v(t)) \tag{32}$$

For the obtained T-S fuzzy model rule, the technique of state feedback is applied to develop the following control rules

$$Model\ rule\ j : If\ z_1(t)\ is\ M_1^j\ and\ z_2(t)\ is\ M_2^j\ and\ z_3(t)\ is\ M_3^j\ and\ z_4(t)\ is\ M_4^j \tag{33}$$
$$Then\ u(t) = K_j x(t), \quad j = 1, \ldots, 16$$

The PDC technique is implemented in this study to find the solution of the system using the obtained T-SFM. Considering Eqs. (29) and (33), PDC technique is applied to design the controller based on T-S fuzzy approach as follow

$$u(t) = \frac{\sum_{j=1}^{16} \omega_j(z(t)) K_j x(t)}{\sum_{j=1}^{16} \omega_j(z(t))} \tag{34}$$

Our developed mode rules in Eq. (29) can be asymptotically stable by a potential positive matrix Q when satisfying the following conditions:

$$Q A_j^T + A_j Q + V_j^T B_j^T + B_j V_j < 0, \quad j = 1, 2, \ldots, 16 \tag{35}$$

$$Q A_j^T + A_j Q + Q A_k^T + A_k Q + V_k^T B_j^T + B_j V_k + V_j^T B_k^T + B_k V_j < 0, \quad j < k \le 16 \tag{36}$$

$$Q = P^{-1} > 0 \tag{37}$$

where $V_j = K_j Q$. Consequently, $K_j$ in Eq. (34) can be calculated by implementing the above LMI conditions.

## 6 A Numerical Example

In this section, simulation tests with MatLab R2017b are implemented to verify the performance of the designed controller for the robot arm model of the parameters that are listed in Tab. 1 with initial arm position and angular velocity $2°$ and $0\ rad/s$, respectively. The control issue of the robot arm system shown in Fig. 1 is assumed to reach the robot arm a desired position by applying a motor torque input. Furthermore, it is assumed that the robot arm is stabilized at a specific angle. The robot arm model is specified in Eq. (13). By using Eq. (35), for 16 rules, we get 16 LMI of the robot arm as follows:

$$QA_1^T + A_1Q + V_1^TB_1^T + B_1V_1 < 0, \tag{38}$$
$$QA_2^T + A_2Q + V_2^TB_2^T + B_2V_2 < 0, \tag{39}$$
$$QA_3^T + A_3Q + V_3^TB_3^T + B_3V_3 < 0, \tag{40}$$
$$QA_4^T + A_4Q + V_4^TB_4^T + B_4V_4 < 0, \tag{41}$$
$$QA_5^T + A_5Q + V_5^TB_5^T + B_5V_5 < 0, \tag{42}$$
$$QA_6^T + A_6Q + V_6^TB_6^T + B_6V_6 < 0, \tag{43}$$
$$QA_7^T + A_7Q + V_7^TB_7^T + B_7V_7 < 0, \tag{44}$$
$$QA_8^T + A_8Q + V_8^TB_8^T + B_8V_8 < 0, \tag{45}$$
$$QA_9^T + A_9Q + V_9^TB_9^T + B_9V_9 < 0, \tag{46}$$
$$QA_{10}^T + A_{10}Q + V_{10}^TB_{10}^T + B_{10}V_{10} < 0, \tag{47}$$
$$QA_{11}^T + A_{11}Q + V_{11}^TB_{11}^T + B_{11}V_{11} < 0, \tag{48}$$
$$QA_{12}^T + A_{12}Q + V_{12}^TB_{12}^T + B_{12}V_{12} < 0, \tag{49}$$
$$QA_{13}^T + A_{13}Q + V_{13}^TB_{13}^T + B_{13}V_{13} < 0, \tag{50}$$
$$QA_{14}^T + A_{14}Q + V_{14}^TB_{14}^T + B_{14}V_{14} < 0, \tag{51}$$
$$QA_{15}^T + A_{15}Q + V_{15}^TB_{15}^T + B_{15}V_{15} < 0, \tag{52}$$
$$QA_{16}^T + A_{16}Q + V_{16}^TB_{16}^T + B_{16}V_{16} < 0, \tag{53}$$

where $Q$ is obtained as explained in Eq. (37). For $j < k \leq 16$, from Eq. (36), we have $QA_j^T + A_jQ + QA_k^T + A_kQ + V_k^TB_j^T + B_jV_k + V_j^TB_k^T + B_kV_j < 0$. Hence, there are 240 LMI can be designed~ as:

- 
  $j = 1, k = 2, j = 1, k = 3, j = 1, k = 4, j = 1, k = 5, j = 1, k = 6, j = 1, k = 7, j = 1, k = 8, j = 1, k = 9, j = 1, k = 10, j = 1, k = 11, j = 1, k$
- 
  $j = 2, k = 3, j = 2, k = 4, j = 2, k = 5, j = 2, k = 6, j = 2, k = 7, j = 2, k = 8, j = 2, k = 9, j = 2, k = 10, j = 2, k = 11, j = 2, k = 12, j = 2, k$
- 
  $j = 3, k = 4, j = 3, k = 5, j = 3, k = 6, j = 3, k = 7, j = 3, k = 8, j = 3, k = 9, j = 3, k = 10, j = 3, k = 11, j = 3, k = 12, j = 3, k = 13, j = 3,$
- 
  $j = 4, k = 5, j = 4, k = 6, j = 4, k = 7, j = 4, k = 8, j = 4, k = 9, j = 4, k = 10, j = 4, k = 11, j = 4, k = 12, j = 4, k = 13, j = 4, k = 14, j = 4$
- 
  $j = 5, k = 6, j = 5, k = 7, j = 5, k = 8, j = 5, k = 9, j = 5, k = 10, j = 5, k = 11, j = 5, k = 12, j = 5, k = 13, j = 5, k = 14, j = 5, k = 15, j =$
- 
  $j = 6, k = 7, j = 6, k = 8, j = 6, k = 9, j = 6, k = 10, j = 6, k = 11, j = 6, k = 12, j = 6, k = 13, j = 6, k = 14, j = 6, k = 15, j = 6, k = 16,$
- $j = 7, k = 8, j = 7, k = 9, j = 7, k = 10, j = 7, k = 11, j = 7, k = 12, j = 7, k = 13, j = 7, k = 14, j = 7, k = 15, j = 7, k = 16,$
- $j = 8, k = 9, j = 8, k = 10, j = 8, k = 11, j = 8, k = 12, j = 8, k = 13, j = 8, k = 14, j = 8, k = 15, j = 8, k = 16,$
- $j = 9, k = 10, j = 9, k = 11, j = 9, k = 12, j = 9, k = 13, j = 9, k = 14, j = 9, k = 15, j = 9, k = 16,$
- $j = 10, k = 11, j = 10, k = 12, j = 10, k = 13, j = 10, k = 14, j = 10, k = 15, j = 10, k = 16,$
- $j = 11, k = 12, j = 11, k = 13, j = 11, k = 14, j = 11, k = 15, j = 11, k = 16,$
- $j = 12, k = 13, j = 12, k = 14, j = 12, k = 15, j = 12, k = 16,$
- $j = 13, k = 14, j = 13, k = 15, j = 13, k = 16,$
- $j = 14, k = 15, j = 14, k = 16,$
- $j = 15, k = 16.$

One of the essential considerations of LMI design is the interconnection of membership $j$ and $k$. For instance, assume $j = 11$ and $k = 12$. Thus, the interaction between the 11$^{th}$ fuzzy rule and the 12$^{th}$ fuzzy rule should be considered. The obtained LMI in this case is

$$QA_{11}^T + A_{11}Q + QA_{12}^T + A_{12}Q + V_{12}^TB_{11}^T + B_{11}V_{12} + V_{11}^TB_{12}^T + B_{12}V_{11} < 0, \tag{54}$$

Referring to Eq. (41), T-SFM controller has been designed by applying PDC technique

$$
\begin{aligned}
u =\ & w_1 (z_1(t)) K_1x(t) + w_2 (z_2(t)) K_2x(t) + w_3 (z_3(t)) K_3x(t) + w_4 (z_4(t)) K_4x(t) \\
& + w_5 (z_5(t)) K_5x(t) + w_6 (z_6(t)) K_6x(t) + w_7 (z_7(t)) K_7x(t) \\
& + w_8 (z_8(t)) K_8x(t) + w_9 (z_9(t)) K_9x(t) + w_{10} (z_{10}(t)) K_{10}x(t) \\
& + w_{11} (z_{11}(t)) K_{11}x(t) + w_{12} (z_{12}(t)) K_{12}x(t) + w_{13} (z_{13}(t)) K_{13}x(t) \\
& + w_{14} (z_{14}(t)) K_{14}x(t) + w_{15} (z_{15}(t)) K_{15}x(t) + w_{16} (z_{16}(t)) K_{16}x(t)
\end{aligned}
\tag{55}
$$

where $K_i = V_iP$, $i = 1, 2, \ldots, 16$. Utilizing both of LMI and YALMIP toolboxes, the values of Q and $V_i$ are obtained. Consequently, $K_i$ is obtained as:

$$K_1 = \begin{bmatrix} 4263 & 3908 & 6811 & 5229 \end{bmatrix}, \quad K_2 = \begin{bmatrix} 1298 & 9084 & 7482 & 9502 \end{bmatrix},$$
$$K_3 = \begin{bmatrix} 6015 & 8918 & 7671 & 4980 \end{bmatrix}, \quad K_4 = \begin{bmatrix} 3173 & 4423 & 7912 & 6245 \end{bmatrix},$$
$$K_5 = \begin{bmatrix} 8903 & 9562 & 3891 & 4291 \end{bmatrix}, \quad K_6 = \begin{bmatrix} 4189 & 6922 & 9826 & 5229 \end{bmatrix},$$
$$K_7 = \begin{bmatrix} 5103 & 7984 & 6712 & 3108 \end{bmatrix}, \quad K_8 = \begin{bmatrix} 6186 & 9612 & 8866 & 5439 \end{bmatrix},$$
$$K_9 = \begin{bmatrix} -7139 & -4700 & -6559 & -5999 \end{bmatrix}, \quad K_{10} = \begin{bmatrix} -5654 & -6357 & -7383 & -9072 \end{bmatrix},$$
$$K_{11} = \begin{bmatrix} -8049 & -4568 & -9641 & -4689 \end{bmatrix}, \quad K_{12} = \begin{bmatrix} -7678 & -6473 & -7213 & -6541 \end{bmatrix},$$
$$K_{13} = \begin{bmatrix} -6984 & -8867 & -6293 & -4796 \end{bmatrix}, \quad K_{14} = \begin{bmatrix} -5180 & -8744 & -1746 & -6244 \end{bmatrix},$$
$$K_{15} = \begin{bmatrix} -3523 & -7650 & -8102 & -9719 \end{bmatrix}, \quad K_{16} = \begin{bmatrix} -5780 & -7217 & -6834 & -7422 \end{bmatrix}.$$

# 7 Results

The simulation results of the proposed model are shown in Figs. 2 and 3. In Fig. 2, the results of the robot arm position and velocity are presented for the equilibrium situation where $x$ and $\dot{x}$ are equal to zero with time above 1.5 $s$. The simulation shows the transient response between $t = 0\,s$ and $t = 1.5\,s$. In which, the arm position is reached to zero at time $0.35\,s$ in a linear way. On the other side, the angular velocity of the arm reached its maximum value 7.8 $rad/s$ in opposite direction at time 0.11 $s$. The dynamic effect of the robot arm is caused by the input torque of the motor presented in Fig. 3. As shown in Fig. 3, to stabilized the robot arm to $0°$ position, the motor should supply a torque 7880 $N.m$ in a short time as an impulse signal. The results of Fig. 3 are useful in determining the suitable features of the potential motor for the proposed robot arm physical parameters.

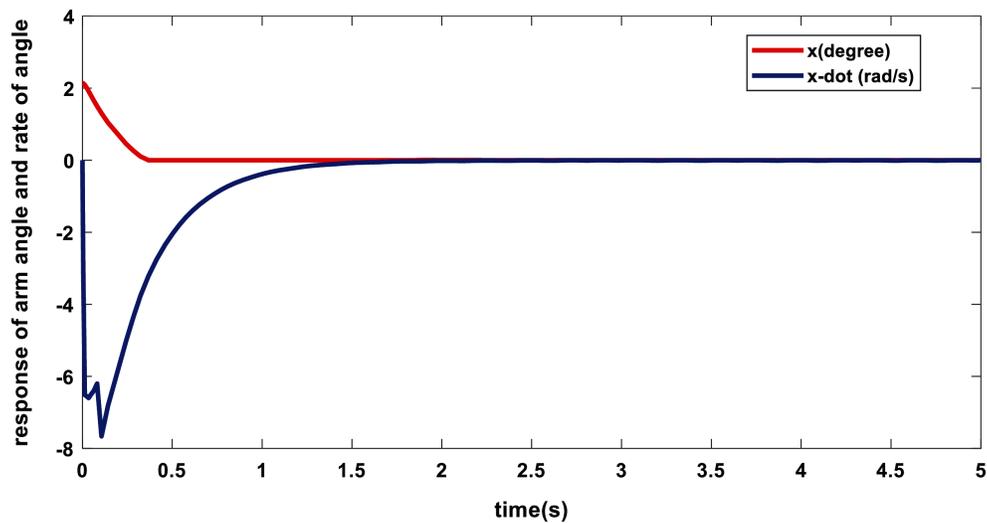

**Figure 2:** States of arm response

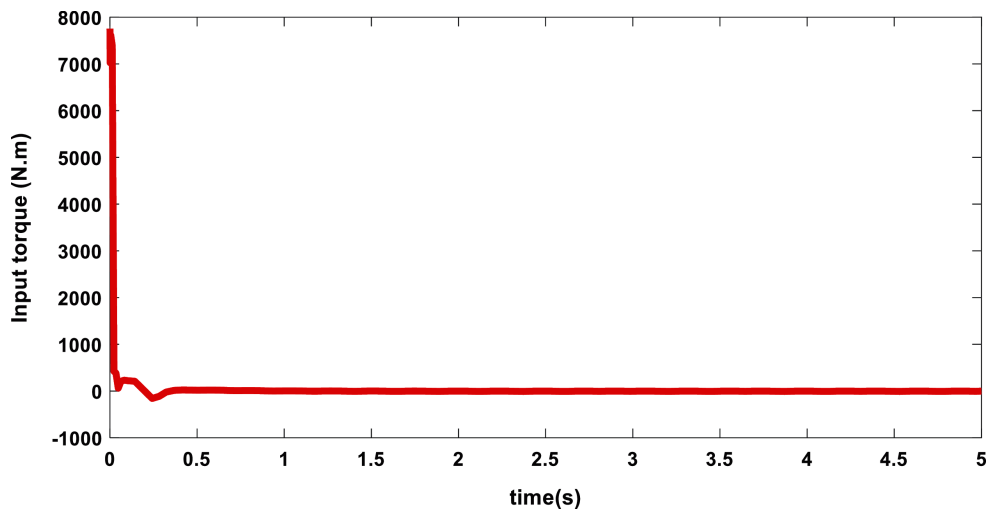

**Figure 3:** Input torque of motor

## 8 Conclusion

In this paper, a new control technique of the rotation motion of a nonlinear robot arm manipulator is introduced. The LMI and PDC control approaches are implemented based on TS-FM of the state space equations. The TS-FM has been developed for linearizing the nonlinear parameters in the state space equations in appropriately selected conditions of the operating points. The prime concept of the designed controller is to deduce all the fuzzy rules using the PDC approach in order to compensate for all rules of the fuzzy model. Furthermore, the linear matrix inequalities (LMIs) are applied to generate adequate cases for approving the stability and control purpose issues. The simulation results demonstrate that the proposed control technique stabilized the system with zero tracking error in less than 1.5 s. This is a suitable control performance for nonlinear robot arm models. The main limitation of this work is the non-utilization of optimization techniques such as genetic algorithms to find the optimum boundaries of membership functions that minimize the tracking errors. This will be addressed in the future by using such approaches to deal with the design membership functions in this application considering their types and boundaries.

**Funding Statement:** The authors received no specific funding for this research study.

**Conflicts of Interest:** The authors declare that they have no conflicts of interest to report regarding the present study.